\documentclass[runningheads]{llncs}
\usepackage{epsfig}

\usepackage{adjustbox}
\usepackage{amsmath,amssymb,amsfonts}
\usepackage{graphicx}
\usepackage{subcaption}
\usepackage{textcomp}
\usepackage{xcolor}
\usepackage{multirow}
\usepackage{float}
\usepackage{mathtools}
\usepackage{booktabs}
\usepackage{hyperref}
\usepackage{rotating}
\usepackage{tikz}
\usepackage{algpseudocode}
\usepackage[misc,geometry]{ifsym}
\usepackage[T1]{fontenc}
\usepackage{graphicx}
\newcommand{\yan}[1]{{\color{black}{{#1}}}}  

\begin{document}
\title{Semantic Image Synthesis for Abdominal CT}
%
%
\author{Yan Zhuang, Benjamin Hou, Tejas Sudharshan Mathai, Pritam Mukherjee, Boah Kim, Ronald M. Summers$^{(\text{\Letter})}$} 
%
\authorrunning{Zhuang et al.}
%
\institute{Imaging Biomarkers and Computer-Aided Diagnosis Laboratory,\\Department of Radiology and Imaging Sciences,\\
National Institutes of Health Clinical Center\\
\email{\{yan.zhuang2,benjamin.hou,tejas.mathai\\pritam.mukherjee,boah.kim,rsummers\}@nih.gov}}
\maketitle              
\begin{abstract}

As a new emerging and promising type of generative models, diffusion models have proven to outperform Generative Adversarial Networks (GANs) in multiple tasks, including image synthesis. In this work, we explore semantic image synthesis for abdominal CT using conditional diffusion models, which can be used for downstream applications such as data augmentation. We systematically evaluated the performance of three diffusion models, as well as to other state-of-the-art GAN-based approaches, and studied the different conditioning scenarios for the semantic mask. Experimental results demonstrated that diffusion models were able to synthesize abdominal CT images with better quality. Additionally, encoding the mask and the input separately is more effective than na\"ive concatenating.

\end{abstract}
\section{Introduction}

Semantic image synthesis aims to generate realistic images from semantic segmentation masks~\cite{park2019SPADE}. This field has a broad range of applications that range from data augmentation and anonymization to image editing \cite{mahapatra2018efficient,lau2018scargan,mok2019learning,hou2023high,shin2018medical,fernandez2021medical}. For instance, Lau $et~al.$ used a conditional Generative Adversarial Network (GAN) and semantic label maps to synthesize  scar tissues in cardiovascular MRI for data augmentation~\cite{lau2018scargan}. Hou $et~al.$ employed a StyleGAN to synthesize pathological retina fundus images from free-hand drawn semantic lesion maps~\cite{hou2023high}. Shin $et~al.$ utilized a conditional GAN to generate abnormal MRI images with brain tumors, and to serve as an anonymization tool~\cite{shin2018medical}. Mahapatra $et~al.$ leveraged a conditional GAN to synthesize chest x-ray images with different disease characteristics by conditioning on lung masks~\cite{mahapatra2018efficient}. Blanco $et~al.$ proposed editing histopathological images by applying a set of arithmetic operations in the GANs' latent space~\cite{fernandez2021medical}. The main objectives of these works were to address the issues of
data scarcity, given the time-consuming, labor-intensive, and extremely costly of obtaining high-quality data and annotations~\cite{greenspan2016guest}. These studies have demonstrated the effectiveness of using synthetic data for downstream tasks, assuming that
GAN-based generative models can generate photo-realistic images.

More recently, several studies have illustrated that diffusion models surpassed GAN-based models in multiple image synthesis tasks~\cite{saharia2022image,dhariwal2021diffusion}, demonstrating an ability to generate realistic and high-fidelity images. Similarly, diffusion models draw an increasing attention in medical imaging, including registration~\cite{kim2022diffusemorph}, segmentation~\cite{wolleb2022diffusionSegmentation}, reconstruction~\cite{chung2022score}, image-to-image translation~\cite{ozbey2023unsupervised}, anomaly detection~\cite{wolleb2022diffusionAnomaly}, and $etc$. Kazerouni $et~al.$ provided a comprehensive review on latest research progress regarding diffusion models for medical imaging~\cite{kazerouni2023diffusion}. Despite being the \textit{de facto} standard for image synthesis, the application of diffusion models for medical semantic image synthesis remain relatively unexplored. To the best of our knowledge, few studies exist for this task. Zhao $et~al.$ employed the Semantic Diffusion Model (SDM) to synthesize pulmonary CT images from segmentation maps for data augmentation~\cite{zhao2023highfidelity}, while Dorjsembe $et~al.$ developed a diffusion model to simulate brain tumors in MRI \cite{dorjsembe2023conditional}.

\begin{figure*}[!htb]
    \centering
    \includegraphics[width=0.8\linewidth]{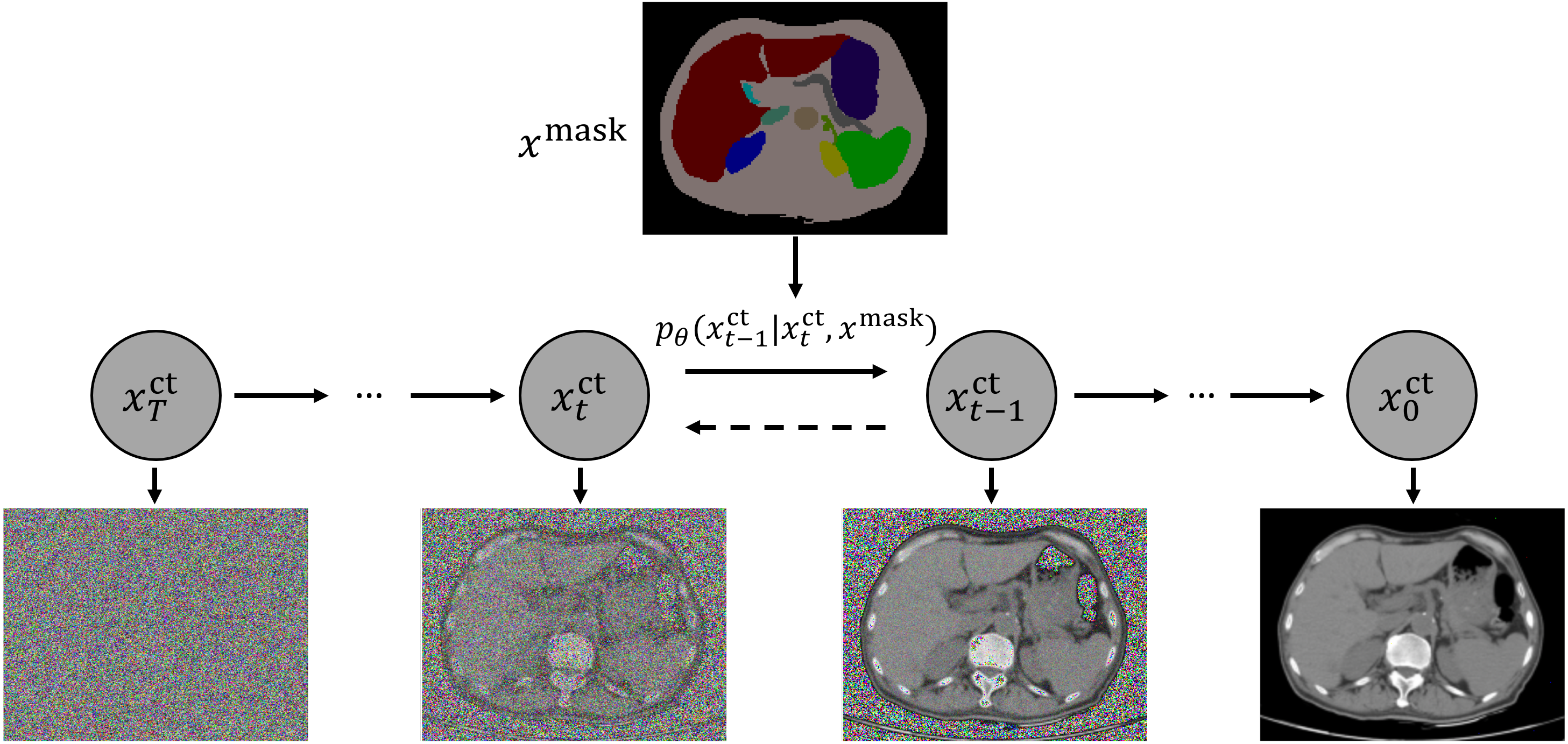}
    \caption{The semantic image synthesis for abdominal CT using a diffusion model. The CT image mask $x^\text{mask}$ guides the diffusion process, which synthesizes a CT image that matches the semantic layout of the mask $x^\text{mask}$. $p_\theta$ is the neural network parameterized by $\theta$, and $x^\text{ct}_t$ is the output of the network at time step $t$. Different colors of $x^\text{mask}$ represent different abdominal organs and structures.}
    \label{fig:pipeline}
\end{figure*}

As prior work primarily used GAN-based models and focused on the head and thorax, our study investigates the use of conditional diffusion models for the semantic medical image synthesis of abdominal CT images as shown in Fig.~\ref{fig:pipeline}. The abdomen is anatomically complex with subtle structures (e.g., lymph nodes) interwoven with large organs (e.g., liver). Consequently, the associated semantic segmentation maps are dense and complex. This complexity presents a far greater challenge when conducting image synthesis for the abdomen. We initially explored different conditioning configurations for diffusion models, such as channel-wise concatenation, encoding the mask in a separated encoder, and the use of auxiliary information. Then we assessed the performance of the conditional diffusion models against GAN-based models in terms of image quality and learned correspondence. Our experimental evaluation demonstrated that encoding the mask and the input enabled the diffusion model to converge earlier and gain improved performance. Moreover, the results showed that conditional diffusion models achieved superior image quality in terms of Fréchet Inception Distance score (FID), Structural Similarity Index Measure (SSIM) and Peak Signal to Noise Ratio (PSNR) scores within a large, publicly available dataset. While diffusion models excelled at learned correspondence in large organs, they were outperformed by GAN-based methods in small structures and organs. Despite this, the conditional diffusion models still yielded promising results. 

Our contributions are two-fold: (1) we demonstrate the effectiveness of diffusion models in the task of semantic image synthesis for abdomen CT and provided a comprehensive comparative evaluation against other State-of-The-Art (SOTA) GAN-based approaches; (2) we empirically show that encoding masks in a separated encoder branch can achieve superior performance, shedding light on finding a more effective way to leverage the semantic mask information.

\section{Method}
\label{sec_method}

Although the process of synthesizing a CT image from a given semantic segmentation mask is a form of conditional image generation, it fundamentally relies on (unconditional) diffusion models. This study focuses on Denoising Diffusion Probabilistic Models (DDPM)~\cite{nichol2021improved}.

\begin{figure*}[!htb]
    \centering
    \begin{minipage}{0.7\textwidth}
        \centering
        \includegraphics[width=0.95\linewidth]{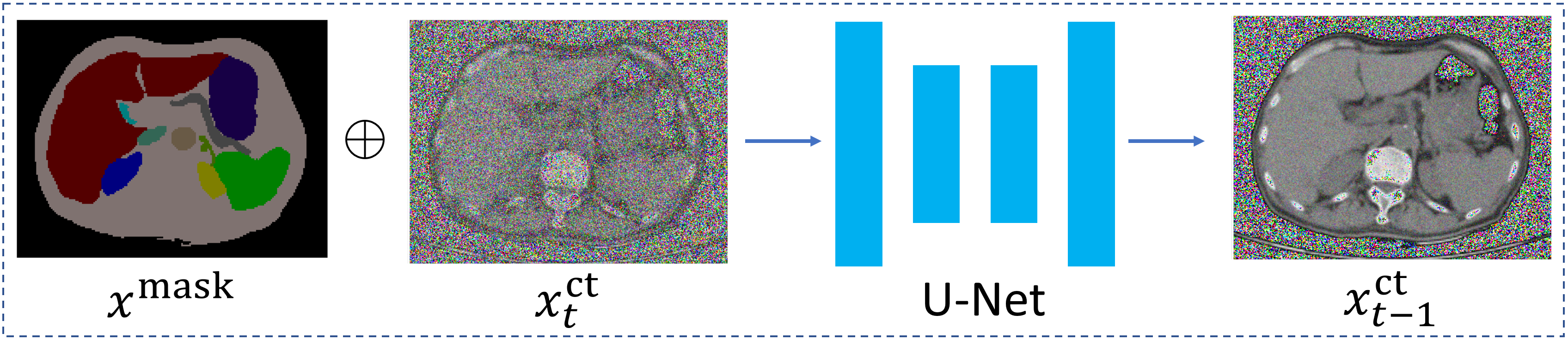}
        \centerline{(a) Conditional DDPM}
        \centerline{}
    \end{minipage}
    \hfill
    \begin{minipage}{0.7\textwidth}
        \centering
        \includegraphics[width=0.95\linewidth]{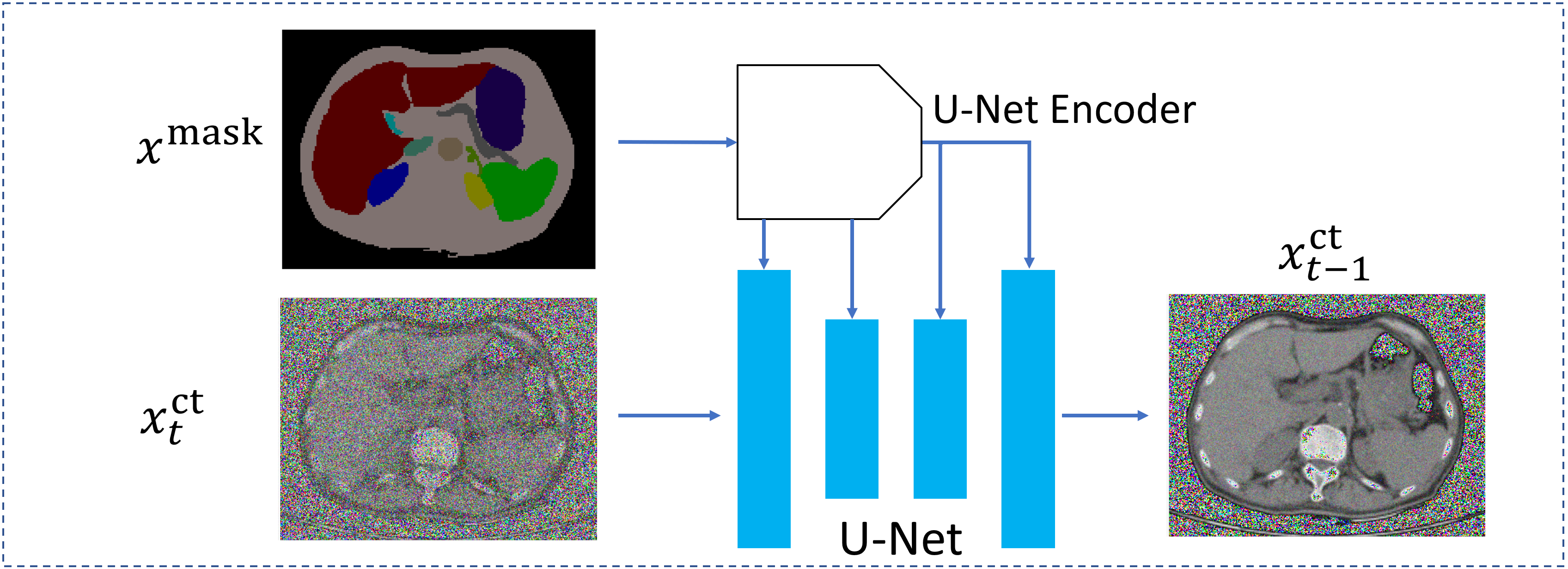}
        \centerline{(b) Mask-guided DDPM}
        \centerline{}
    \end{minipage}
    \hfill
    \begin{minipage}{0.7\textwidth}
        \centering
        \includegraphics[width=0.95\linewidth]{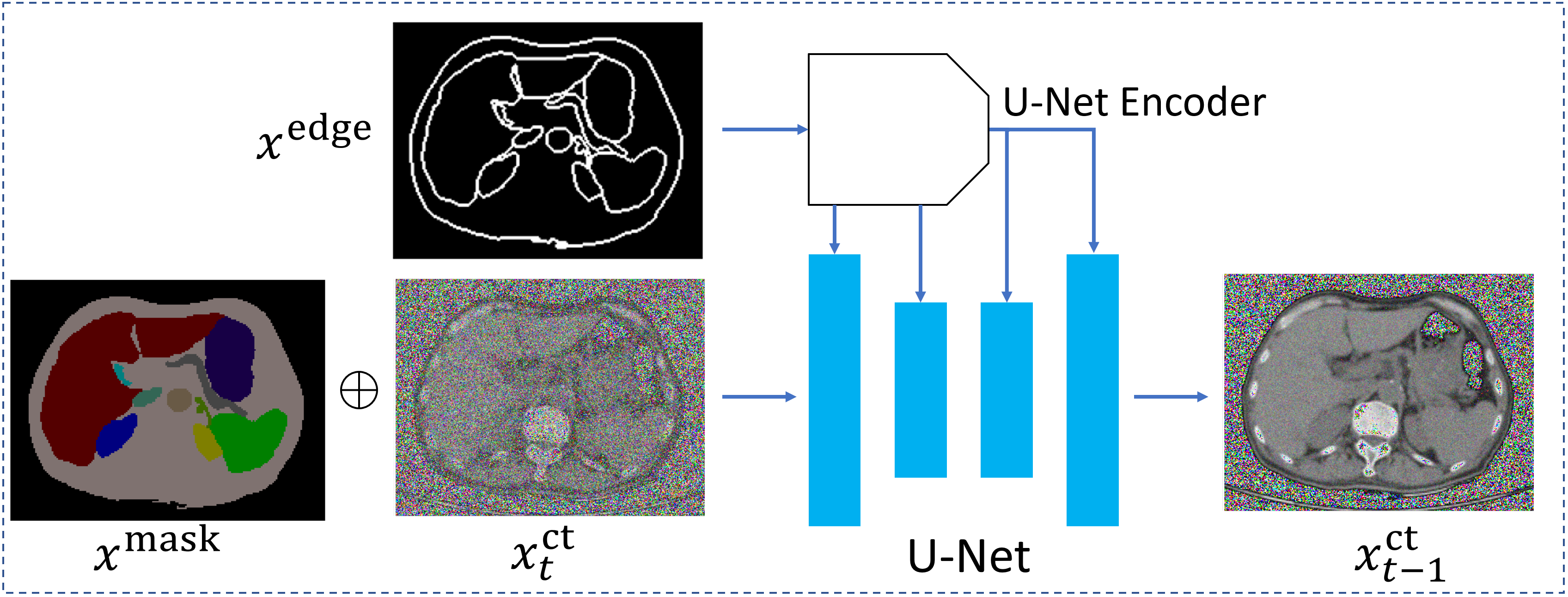}
        \centerline{(c) Edge-guided DDPM}
        \centerline{}
    \end{minipage}

\caption{Different conditions of diffusion models for semantic image synthesis of abdominal CT images: (a) channel-wise concatenating, denoted as ``conditional DDPM''; (b) mask guidance where the conditioning mask is encoded in a U-Net encoder, denoted as ``mask-guided DDPM''; (c) using other auxiliary information, e.g., semantic edge map, denoted as ``edge-guided DDPM''.}
\label{fig2_arch} 
\end{figure*}

The DDPM model consists of a forward diffusion process and a reverse diffusion process. The forward process progressively transforms a clean image into an image with isotropic Gaussian noise. Mathematically, considering a clean image sample $x^{\text{ct}}_0$ and a set of time steps $\{1, \cdots,t, \cdots, T\}$, Gaussian noise is progressively added to the image at time step $t$ by:
\begin{equation}
    q(x^{\text{ct}}_t|x^{\text{ct}}_{t-1}) = \mathcal{N}(x^{\text{ct}}_t; \sqrt{1- \beta_t} x^{\text{ct}}_{t-1}, \beta_t \bf{I}),
\label{eq.forward}
\end{equation}
where $\beta_t$ is the scheduled variance. Then, using Markov chain rule, the forward process of $x^{\text{ct}}_t$ from $x^{\text{ct}}_0$ can be formulated by:
\begin{equation}
    q(x^{\text{ct}}_t|x^{\text{ct}}_0) = \mathcal{N}(x^{\text{ct}}_t; \sqrt{\bar{\alpha}_t} x^{\text{ct}}_0, (1-\bar{\alpha}) \textbf{I}),
\end{equation}
where $\alpha_t = 1- \beta_t$ and $ \bar{\alpha}_t = \prod^t_{s=1} \alpha_s$. Accordingly, for $\epsilon \in \mathcal{N}(\text{0}, \bf{I})$, a noisy image $x_t$ can be expressed in terms of $x_0$ in a closed form:
\begin{equation}
    x^{\text{ct}}_t = \sqrt{\bar{\alpha}_t}x^{\text{ct}}_0 + \sqrt{1-\bar{\alpha}_t}\epsilon.
\end{equation}

On the other hand, the reverse diffusion process gradually removes Gaussian noise by approximating $q(x_{t-1}|x_t)$ through a neural network $p_\theta$ parameterized by $\theta$,
\begin{equation}
    p_\theta(x^{\text{ct}}_{t-1}|x^{\text{ct}}_t) = \mathcal{N}(x^{\text{ct}}_{t-1}; \mu_\theta(x^{\text{ct}}_t,t),\Sigma_\theta(x^{\text{ct}}_t, t)),
\end{equation}
where $\mu_\theta$ and $\Sigma_\theta$ are predicted mean and variance. Thus, the image sample $x_{t-1}$ at time step $t-1$ can be predicted as:
\begin{equation}
    x^{\text{ct}}_{t-1} = \frac{1}{\sqrt{\alpha_t}}\left(x^{\text{ct}}_t-\frac{\beta_t}{\sqrt{1-\bar{\alpha}}_t}\epsilon_\theta(x^{\text{ct}}_t,t)\right) + \sigma_t\bf{z},
\label{eq.sampling}
\end{equation}
where $\epsilon_\theta$ is the trained U-Net, $\sigma_t$ is the learned variance, and $\bf{z} \in \mathcal{N}(\text{0}, \bf{I})$. A detailed formulation of DDPM can be found in \cite{nichol2021improved}. 

The aforementioned formulation of DDPM is an unconditional image synthesis process, meaning that the synthetic CT images are generated from random anatomic locations. However, our goal is a conditional image synthesis process. The aim is to generate the CT images in such a way that the synthetic CT images preserve the same semantic layout as the given input CT masks. The input CT image mask, denoted by $x^\text{mask}$, should guide the diffusion process and synthesize an image that matches the semantic layout of the mask. In this pilot work, to assess the effectiveness of various conditioning methods, we have presented three different conditioning scenarios: 
(1) channel-wise concatenating, denoted as ``conditional DDPM''; 
(2) mask guidance where the conditioning mask is encoded in a separated network branch, denoted as ``mask-guided DDPM''; 
(3) using other auxiliary information, e.g., semantic edge map, denoted as ``edge-guided DDPM''.

\paragraph{Conditional DDPM.}
In this method, the idea was to concatenate the mask $x^\text{mask}$ together with the input image $x_t$ in an additional input channel. \yan{The network architecture was as the same as the DDPM model, as shown in Fig.~\ref{fig2_arch}(a)}. Then, in this case, eq.~(\ref{eq.sampling}) became: 
\begin{equation}
    x^{\text{ct}}_{t-1} = \frac{1}{\sqrt{\alpha_t}}\left(x^{\text{ct}}_t-\frac{\beta_t}{\sqrt{1-\bar{\alpha}}_t}\epsilon_\theta(x^{\text{ct}}_t\oplus x^\text{mask}, t)\right) + \sigma_t\bf{z},
\end{equation}
where $x^{\text{ct}}_t\oplus x^\text{mask}$ is the channel-wise concatenation of the input image $x_t$ and the given mask $x^\text{mask}$. 

\paragraph{Mask-guided DDPM.}
The second strategy was to encode the mask separately by employing another U-Net-like encoder, and injecting the encoding information directly into the main U-Net branch. \yan{More specifically, feature maps from the convolutional layers before each downsampling layer of the U-Net-like encoder were concatenated to the corresponding feature maps of the main U-Net branch encoder and decoder, as shown in Fig.~\ref{fig2_arch}(b)}. This idea was similar to SPADE~\cite{park2019SPADE}. Then in this case, eq.~(\ref{eq.sampling}) became:
\begin{equation}
    x^{\text{ct}}_{t-1} = \frac{1}{\sqrt{\alpha_t}}\left(x^{\text{ct}}_{t}-\frac{\beta_t}{\sqrt{1-\bar{\alpha}}_t}\epsilon_\theta(x^{\text{ct}}_t, x^\text{mask}, t)\right) + \sigma_t\bf{z}.
\end{equation}

\paragraph{Edge-guided DDPM.}
Finally, rather than using only the semantic mask, we can leverage the semantic edge map, $e.g.,$ $x^\text{edge}$, as the auxiliary information to guide the diffusion process. \yan{The network architecture was a combination of ``conditional DDPM'' and ``Mask-guided DDPM'', as shown in Fig.~\ref{fig2_arch}(c).} Then in this case, eq.~(\ref{eq.sampling}) became: 
\begin{equation}
    x^{\text{ct}}_{t-1} = \frac{1}{\sqrt{\alpha_t}}\left(x^\text{ct}_{t}-\frac{\beta_t}{\sqrt{1-\bar{\alpha}}_t}\epsilon_\theta(x^{\text{ct}}_{t}\oplus x^\text{mask}, x^\text{edge},t)\right) + \sigma_t\bf{z}.
\end{equation}

Our implementation was based on~\cite{nichol2021improved}. Specifically, we set the time step $T=1000$ and we used a linear noisy scheduler. The network used a ResNet backbone~\cite{he2016deep}. A hybrid loss function consisting of a $L_2$ loss item and variational lower-bound loss item was utilized to train the network. The optimizer was Adam with a learning rate of $1e^{-4}$. We trained the models for 150k iterations using a batch size of 16.

\section{Experiments}
\label{sec_exp_setup}

\noindent
\textbf{Dataset.} We used the training set of AMOS22~\cite{ji2022amos} CT subset to train all models. This training dataset consisted of 200 CT volumes of the abdomen from 200 different patients. The CT data was collected from multiple medical centers using different scanners, with detailed image acquisition information available in \cite{ji2022amos}. The dataset contained voxel-level annotations of 15 abdominal organs and structures: spleen, right kidney, left kidney, gallbladder, esophagus, liver, stomach, aorta, inferior vena cava, pancreas, right adrenal gland, left adrenal gland, duodenum, bladder, prostate/uterus. We added an additional ``body'' class to include the remaining structures beyond these 15 organs, and this was obtained through thresholding and morphological operations. The testing split comprised 50 CT volumes from 50 subjects that were taken from the AMOS22 CT validation set. We pre-processed the CT images by applying a windowing operation with a level of 40 and a width of 400. Each 2D slice in the 3D CT volume was extracted, normalized to the range of [0, 1], and resized to 256 $\times$ 256 pixels. This resulted in 26,069 images from 200 subjects for the training set, and 6559 images from 50 subjects for the testing set.

\noindent
\textbf{Baseline Comparisons.} To comprehensively evaluate the performance of the proposed diffusion-based approaches, we compared them with several SOTA semantic image synthesis methods. These included GAN-based methods, such as SPADE~\cite{park2019SPADE}, OASIS~\cite{schnfeld2021you}, Pix2Pix~\cite{pix2pix2017}, as well as an existing diffusion-based approach SDM~\cite{wang2022semantic}. All comparative methods were implemented in PyTorch. The learning rates for SPADE and Pix2Pix were set at  $5e^{-4}$. For OASIS, the learning rates were set at $4e^{-4}$ for the Discriminator and $1e^{-4}$ for the Generator. We used the Adam optimizer to train all models, for 300 epochs, with the most recent checkpoint used for evaluation. For the SDM model, we adhered to the same training scheme used for DDPM-based models. 

\noindent
\textbf{Evaluation metrics.} We evaluated the performance based on both visual quality and learned organ correspondence. To assess visual quality, we used FID, SSIM, and PSNR. For assessing learned organ correspondence, we utilized an off-the-shelf, CT-only multi-organ segmentation network named TotalSegmentator (TS) \cite{wasserthal2022totalsegmentator}. This network was used to predict segmentation masks from synthetic CT volumes. Subsequently, Dice coefficients (DSC) were computed to compare these predicted masks with the ground-truth annotations.

\section{Results and Discussion}
\label{sec_result}

\begin{figure*}[!htb]
    \centering
    \begin{minipage}{0.75\textwidth}
        \centering
        \includegraphics[width=1\linewidth]{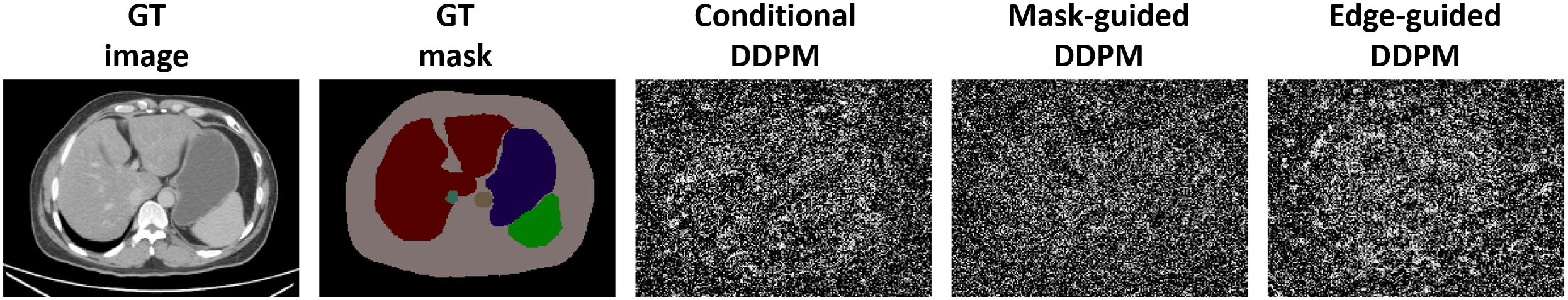}
    \end{minipage}
    \hfill
    \begin{minipage}{0.75\textwidth}
        \centering
        \includegraphics[width=1\linewidth]{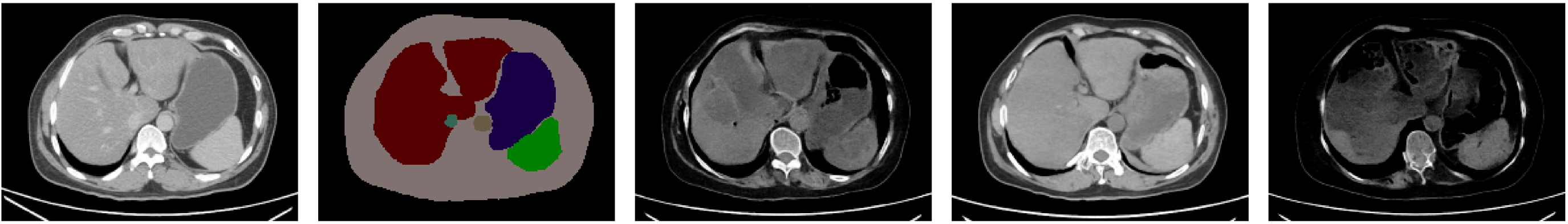}
    \end{minipage}
    \hfill
    \begin{minipage}{0.75\textwidth}
        \centering
        \includegraphics[width=1\linewidth]{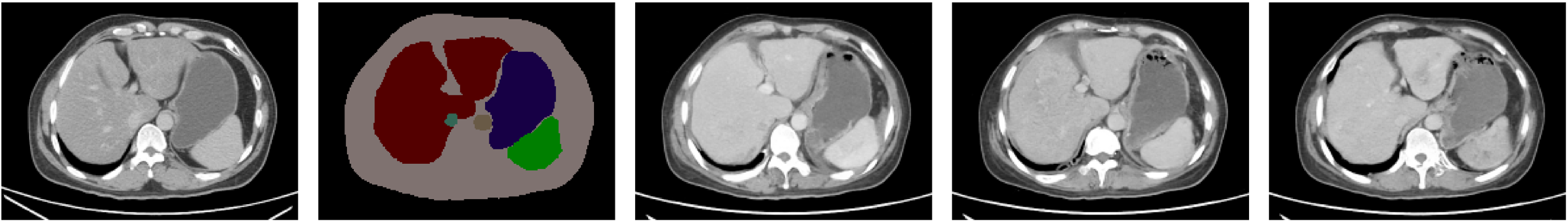}      
    \end{minipage}
    \hfill
    \begin{minipage}{0.75\textwidth}
        \centering
        \includegraphics[width=1\linewidth]{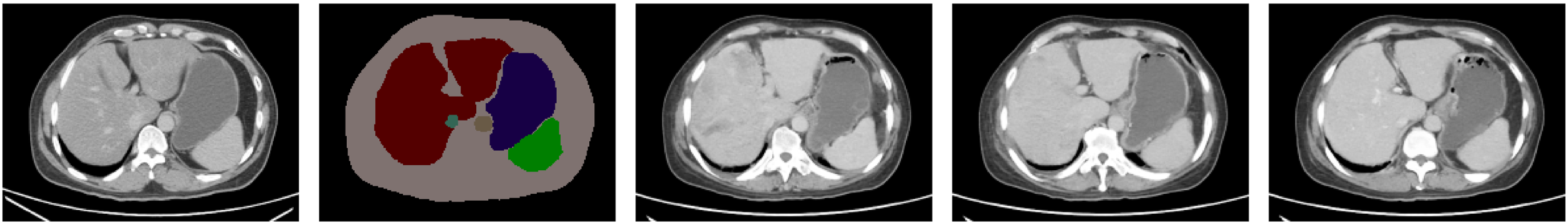}
    \end{minipage}
    
\caption{Sampling results for three conditioning scenarios after 10k, 50k, 100k, and 150k training iterations. The first row shows the results after 10k training iterations; the second row shows results after 50k training iterations; the third row shows the results after 100k training iterations; the fourth row shows the results after 150k training iterations. The color map for different organs: liver (dark red), stomach (indigo), spleen (green), aorta (light brown), and inferior vena cava (aqua blue green).}
\label{Fig_three_models} 
\end{figure*}

\noindent
\textbf{Training iteration study.}
We initially evaluated the three different conditioning strategies after 50k, 100k, and 150k training iterations. Table~\ref{tab1_diff_cond} presents the numerical results. The overall trend indicated that the performance of all three proposed models converged after 150k training iterations. The mask-guided DDPM model outperformed the others by a small margin in most metrics at the 150k iteration mark. However, at earlier stages of training, specifically after 50k training iterations, the mask-guided DDPM model surpassed the conditional DDPM and edge-guided DDPM models, implying an earlier convergence. Furthermore, we observed that using auxiliary edge-map information did not improve performance. Fig.~\ref{Fig_three_models} visualizes sample images after 10k, 50k, 100k, and 150k training iterations, respectively. 

\begin{table}[!htbp]
  \centering
    \caption{FID, PSNR, SSIM, and DSC scores. The highest performance in each column is highlighted for different iterations setups.}
    \resizebox{\linewidth}{!}{
    \begin{tabular}{c|c|ccc|cccccccccccccc}
    \toprule
    \multirow{2}[2]{*}{Iter.} & \multirow{2}[2]{*}{Methods} &       &       &       & \multicolumn{14}{c}{DSC (\%)$\uparrow$} \\
          &       & {FID$\downarrow$} & {PSNR$\uparrow$} & {SSIM$\uparrow$} & {Sple.} & {Liv.} & {Kid\_l} & {Kid\_r} & {Panc.} & {Stom.} & {Aorta} & {Gall\_bld} & {Espo.} & {Adr\_r.} & {Adr\_l.} & {Duod.} & {Cava.} & {Bladder} \\
    \midrule
    \multirow{3}[2]{*}{50k} & {Conditional DDPM} & {30.57} & {14.17} & {0.589} & {77.6} & {75.0} & {\textbf{91.1}} & {87.8} & {57.3} & {60.6} & {\textbf{90.4}} & {30.4} & {66.4} & {\textbf{54.1}} & {53.8} & {\textbf{61.7}} & {\textbf{77.7}} & {\textbf{64.4}} \\
          & {Mask-guided DDPM} & {\textbf{19.06}} & {\textbf{14.58}} & {\textbf{0.603}} & {\textbf{83.9}} & {\textbf{84.8}} & {90.3} & {\textbf{90.2}} & {\textbf{62.1}} & {\textbf{73.4}} & {89.2} & {\textbf{40.0}} & {\textbf{73.2}} & {53.2} & {56.6} & {54.1} & {75.2} & {57.0} \\
          & {Edge-guided DDPM} & {35.97} & {13.43} & {0.576} & {56.4} & {56.7} & {86.6} & {82.5} & {51.2} & {52.2} & {86.8} & {10.6} & {58.3} & {52.6} & {\textbf{58.2}} & {53.6} & {73.7} & {61.2} \\
    \midrule
    \multirow{3}[2]{*}{100k} & {Conditional DDPM} & {11.27} & {16.07} & {\textbf{0.643}} & {\textbf{93.8}} & {95.3} & {93.9} & {92.5} & {73.8} & {85.0} & {91.1} & {64.5} & {76.6} & {\textbf{66.5}} & {62.8} & {65.1} & {81.0} & {70.8} \\
          & {Mask-guided DDPM} & {10.89} & {16.10} & {0.642} & {\textbf{93.8}} & {\textbf{95.8}} & {93.9} & {93.1} & {\textbf{75.3}} & {\textbf{86.9}} & {\textbf{91.5}} & {63.9} & {\textbf{79.6}} & {65.9} & {\textbf{68.5}} & {\textbf{68.0}} & {\textbf{82.5}} & {\textbf{71.6}} \\
          & {Edge-guided DDPM} & {\textbf{10.32}} & {\textbf{16.14}} & {0.644} & {93.6} & {95.1} & {\textbf{94.1}} & {\textbf{93.4}} & {73.8} & {85.1} & {90.8} & {\textbf{65.0}} & {77.3} & {64.9} & {64.1} & {64.5} & {80.6} & {71.0} \\
    \midrule
    \multirow{3}[2]{*}{150k} & {Conditional DDPM} & {\textbf{10.56}} & {16.26} & {\textbf{0.646}} & {\textbf{94.0}} & {\textbf{95.6}} & {\textbf{93.9}} & {\textbf{91.2}} & {\textbf{76.3}} & {86.4} & {90.8} & {64.0} & {78.2} & {\textbf{67.2}} & {\textbf{66.0}} & {\textbf{65.6}} & {80.9} & {70.0} \\
          & {Mask-guided DDPM} & {10.58} & {\textbf{16.28}} & {\textbf{0.646}} & {93.9} & {\textbf{95.6}} & {\textbf{93.9}} & {90.9} & {75.6} & {\textbf{87.1}} & {\textbf{91.3}} & {\textbf{66.0}} & {\textbf{79.5}} & {\textbf{67.2}} & {65.1} & {\textbf{65.6}} & {\textbf{81.3}} & {\textbf{70.7}} \\
          & {Edge-guided DDPM} & {10.64} & {16.20} & {\textbf{0.646}} & {93.5} & {95.4} & {93.8} & {92.8} & {75.0} & {86.7} & {90.3} & {64.5} & {78.1} & {65.4} & {64.1} & {65.3} & {79.6} & {69.3} \\
    \bottomrule
    \end{tabular}%
    }
  \label{tab1_diff_cond}%
\end{table}%

\begin{figure*}[!htb]
    \centering
    \begin{minipage}{1\textwidth}
        \centering
        \includegraphics[width=1\linewidth]{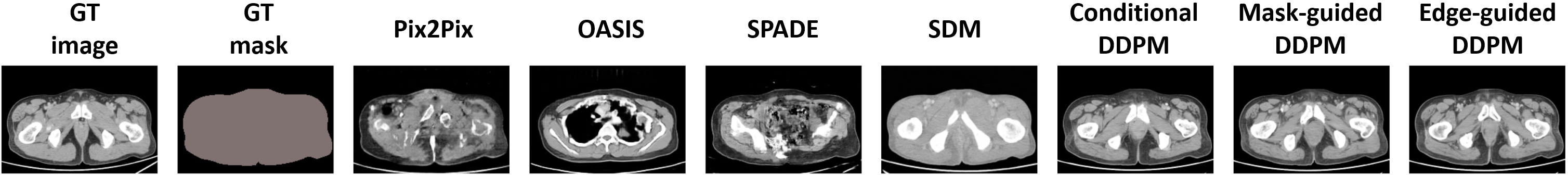}
    \end{minipage}
    \hfill
    \begin{minipage}{1\textwidth}
        \centering
        \includegraphics[width=1\linewidth]{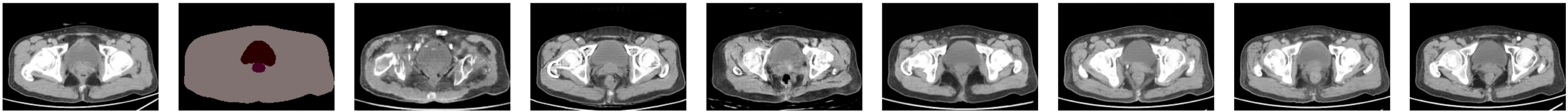}
    \end{minipage}
    \hfill
    \begin{minipage}{1\textwidth}
        \centering
        \includegraphics[width=1\linewidth]{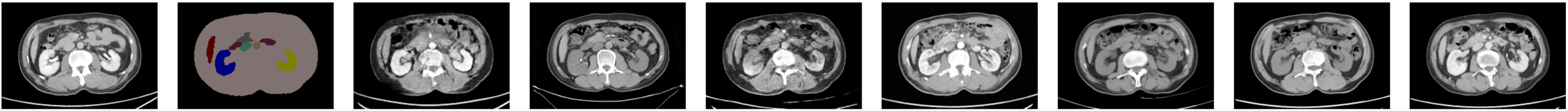}
    \end{minipage}
    \hfill
    \begin{minipage}{1\textwidth}
        \centering
        \includegraphics[width=1\linewidth]{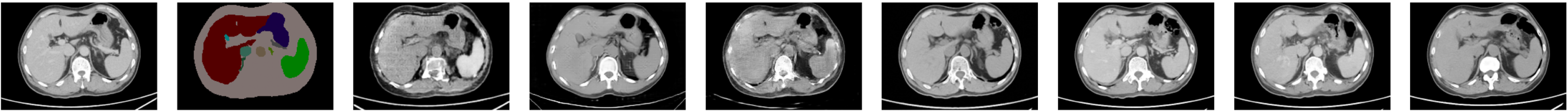}
    \end{minipage}
    \hfill
    \begin{minipage}{1\textwidth}
        \centering
        \includegraphics[width=1\linewidth]{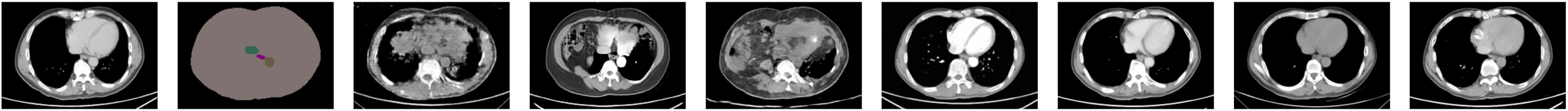}
    \end{minipage}
\caption{Results from different semantic image synthesis methods. The color map for different organs: body (beige), spleen (green), liver (dark red), right kidney (blue), left kidney (yellow), stomach (indigo), aorta (light brown), duodenum (light purple), pancreas (gray), right/left adrenal gland (dark/light green), inferior vena cava (aqua blue green), bladder (shallow brown), prostate (purple).}
\label{Fig_results_methods} 
\end{figure*}

\noindent
\textbf{Comparison study.} We carried out a quantitative evaluation of the diffusion models against other SOTA algorithms such as Pix2Pix, OASIS, SPADE, and SDM methods. In terms of image quality metrics such as FID, PSNR, and SSIM, the diffusion models  outperformed non-diffusion-based methods. However, in terms of learned correspondence metrics like DSC, diffusion models surpassed other models for larger organs such as the liver, spleen, and kidneys. The OASIS method achieved superior performance for relatively small organs and structures like gallbladder and left adrenal gland. This may be because OASIS was good at synthesizing the clear boundary between small organs and the background, resulting in better segmentation results by TS and thus having higher DSC scores. Fig.~\ref{Fig_results_methods} presents multiple results ranging from the lower to the upper abdomen, from different methods. It is worth noting that from the top row of Fig.~\ref{Fig_results_methods} GAN-based methods struggled to synthesize images when the number of mask classes was sparse. For example, Pix2Pix and SPADE were unable to generate realistic images. OASIS generated an image from the upper abdomen, which was inconsistent with the location of the given mask. The bottom row illustrated the same trend. GAN-based models failed to synthesize the context information within the body mask, for example, the heart and lung. On the contrary, diffusion models including the SDM model succeed to generate reasonable images based on the given masks. One explanation was that diffusion models were more effective when the number of masks decreased and the corresponding supervision became sparser.

\begin{table*}[!htbp]
  \centering
  \caption{FID, PSNR, SSIM, and DSC scores for comparable methods. The highest performance in each column is highlighted.}
    \begin{adjustbox}{max width=\textwidth}
    \begin{tabular}{c|ccc|cccccccccccccc}
    \toprule
    \multirow{2}[2]{*}{Methods} &       &       &       & \multicolumn{14}{c}{DSC(\%)$\uparrow$} \\
          & FID $\downarrow$   & PSNR $\uparrow$  & SSIM $\uparrow$  & Sple. & Liv.  & Kid\_l & Kid\_r & Panc. & Stom. & Aorta & Gall\_bld. & Espo. & Adr\_r. & Adr\_l. & Duod. & Cava. & Bladder \\
    \midrule
    {Pix2Pix} & {78.86} & {15.04} & {0.561} & {79.3} & {94.7} & {\textbf{94.5}} & {92.0} & {68.8} & {86.5} & {80.8} & {53.5} & {67.9} & {58.4} & {61.8} & {61.0} & {75.9} & {56.0} \\
    {OASIS} & {43.57} & {14.75} & {0.560} & {91.8} & {93.5} & {92.0} & {88.1} & {73.4} & {\textbf{88.9}} & {88.6} & {\textbf{78.9}} & {\textbf{80.1}} & {\textbf{69.0}} & {\textbf{75.6}} & {\textbf{71.7}} & {\textbf{86.3}} & {\textbf{74.5}} \\
    {SPADE} & {60.22} & {15.27} & {0.594} & {92.6} & {95.4} & {92.7} & {91.2} & {66.6} & {86.5} & {86.2} & {44.8} & {65.6} & {61.7} & {59.3} & {59.5} & {79.1} & {63.9} \\
    \midrule
    {SDM} & {12.68} & {15.12} & {0.607} & {91.9} & {94.3} & {93.5} & \textbf{93.2} & {\textbf{79.3}} & {87.8} & {89.0} & {71.9} & {77.5} & {69.7} & {69.6} & {66.7} & {81.5} & {66.2} \\
    {Conditional DDPM} & {\textbf{10.56}} & {16.26} & {\textbf{0.646}} & {\textbf{94.0}} & {\textbf{95.6}} & {93.9} & {91.2} & {76.3} & {86.4} & {90.8} & {64.0} & {78.2} & {67.2} & {66.0} & {65.6} & {80.9} & {70.0} \\
    {Mask-guided DDPM} & {10.58} & {\textbf{16.28}} & {\textbf{0.646}} & {93.9} & {95.6} & {93.9} & {90.9} & {75.6} & {87.1} & {\textbf{91.3}} & {66.0} & {79.5} & {67.2} & {65.1} & {65.6} & {81.3} & {70.7} \\
    {Edge-guided DDPM} & {10.64} & {16.20} & {\textbf{0.646}} & {93.9} & {95.4} & {93.8} & {92.8} & {75.0} & {86.7} & {90.3} & {64.5} & {78.1} & {65.4} & {64.1} & {65.3} & {79.6} & {69.3} \\
    \bottomrule
    \end{tabular}%
    \end{adjustbox}
  \label{tab:addlabel}%
\end{table*}%

\noindent
\textbf{Future work.} One important application of generative models in medical imaging is to synthesize images for data augmentation. In the future work, we will use diffusion models as a data augmentation strategy and evaluate it in downstream segmentation, classification, or detection tasks. Compared with GAN-based generative models, the major limitation of diffusion models is that sampling procedures are more time-consuming and computationally expensive~\cite{kazerouni2023diffusion}. Nevertheless, multiple recent works successfully showed that using a reduced number of denoising steps was able to obtain high-quality samples, leading to faster inference procedures~\cite{xiao2022tackling}\cite{song2021denoising}. Therefore, by incorporating these techniques, we will investigate the role of conditional masks in fast sampling for synthesizing abdominal CT images.

\section{Conclusion}

In this work, we systematically investigated diffusion models for image synthesis for abdominal CT. Experimental results demonstrated that diffusion models outperformed GAN-based approaches in several setups. In addition, we also showed that disentangling mask and input contributed to performance improvement for diffusion models.

\section*{Acknowledgments}

\noindent
This work was supported by the Intramural Research Program of the National Institutes of Health (NIH) Clinical Center (project number 1Z01 CL040004). This work utilized the computational resources of the NIH HPC Biowulf cluster.

%
%
%
\bibliographystyle{splncs04}
\bibliography{mybibliography}

\end{document}